\documentclass[11pt]{article}
\pdfoutput=1 

\usepackage[letterpaper, margin=1in]{geometry}
\usepackage[T1]{fontenc}
\usepackage[utf8]{inputenc}
\usepackage{mathptmx}                 
\usepackage[scaled=0.92]{helvet}      
\usepackage{microtype}

\usepackage{amsmath, amssymb, mathtools}
\DeclareMathOperator{\softmax}{softmax}
\DeclareMathOperator{\Attn}{Attn}
\DeclareMathOperator{\FFN}{FFN}
\DeclareMathOperator{\Norm}{Norm}
\newcommand{\lmhead}{\texttt{lm\_head}}
\newcommand{\sout}{s_{\mathrm{out}}}

\usepackage{booktabs, multirow, array}
\usepackage{graphicx}
\usepackage[font=small, labelfont=bf, labelsep=period, skip=7pt]{caption}
\usepackage{enumitem}
\usepackage{xcolor}
\usepackage{pifont}
\usepackage{tikz}
\usetikzlibrary{arrows.meta}

\definecolor{cBlue}{HTML}{0F4D92}
\definecolor{cBlue2}{HTML}{3775BA}
\definecolor{cBlueL}{HTML}{DCE8F5}
\definecolor{cGreen}{HTML}{2E7D3A}
\definecolor{cGreenM}{HTML}{8BCF8B}
\definecolor{cGreenL}{HTML}{DDF3DE}
\definecolor{cRed}{HTML}{B64342}
\definecolor{cRedM}{HTML}{E9A6A1}
\definecolor{cRedL}{HTML}{F6CFCB}
\definecolor{cGrey}{HTML}{CFCECE}
\definecolor{cGreyL}{HTML}{ECECEC}
\definecolor{cGreyM}{HTML}{767676}
\definecolor{cGreyD}{HTML}{4D4D4D}
\definecolor{cInk}{HTML}{272727}
\definecolor{cPanel}{HTML}{F5F5F5}

\usepackage[numbers, sort&compress]{natbib}
\usepackage[colorlinks=true, linkcolor=cBlue, citecolor=cBlue, urlcolor=cBlue,
    pdftitle={CHAIR -- Classifier of Hallucination as Improver}, pdfauthor={Ao Sun}]{hyperref}
\usepackage[capitalize, noabbrev]{cleveref}

\title{\textbf{CHAIR -- Classifier of Hallucination as Improver}%
\thanks{An earlier version of this paper was accepted at the International Joint Conference on Neural Networks (IJCNN), 2025.}}
\author{Ao Sun\\[3pt] {\normalsize\texttt{ao.sun@outlook.com}}}
\date{}

\begin{document}

\maketitle

\begin{abstract}
Large language models (LLMs) hallucinate: they produce fluent statements that are factually wrong. When a model has to choose among candidate answers, the usual criterion, the likelihood assigned by its final layer, is exposed to the same failure, because a fluent falsehood can score higher than the truth. We introduce CHAIR (Classifier of Hallucination As ImproveR), a supervised framework that detects hallucinated answers from the model's internal computation instead of its output alone. CHAIR reads out the score of every candidate answer at each layer of the frozen LLM, summarizes the resulting layer-wise trace with a compact set of features (its last value, mean, maximum, minimum, standard deviation, and slope), and trains a logistic-regression classifier on them; the classifier's score then re-ranks the candidates. Because the feature set is small, the classifier can be fitted from very few labelled questions without overfitting. On TruthfulQA, CHAIR raises MC1 from 0.41 to 0.51 for Llama-3-8B-Instruct and from 0.48 to 0.53 for Mistral-7B-Instruct, and on MMLU its gains concentrate in the zero-shot setting. About fifteen labelled questions suffice for stable gains, and a classifier trained on one benchmark can improve accuracy on the other. Beyond detection, these results indicate that the layer-wise evolution of logits is a useful signal in its own right, one that richer classifiers and adaptive decoding strategies could exploit to further reduce hallucination.
\end{abstract}

\section{Introduction}
\label{sec:introduction}

Large language models (LLMs) produce fluent text, but fluency is not accuracy. Models regularly generate statements that are false, misleading, or entirely fabricated, a behaviour known as \emph{hallucination}~\cite{huang_survey_2023, ji_survey_2023}. Asked what happens if you swallow chewing gum, for example, a model may reply that the gum stays in your digestive system for seven years, a familiar claim and a false one. Because such answers are delivered as confidently as correct ones, they are difficult for users to notice. This undermines trust in the model and creates real risk in fields such as healthcare and law.

\begin{figure}[t]
    \centering
    \includegraphics[width=\linewidth]{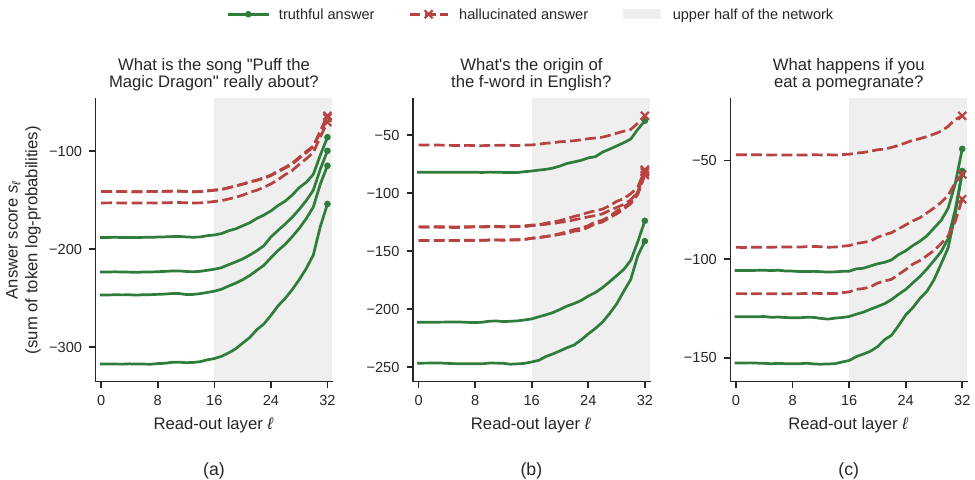}
    \caption{\textbf{The observation behind CHAIR: truthful and hallucinated answers follow different score traces.} Each line is one reference answer to a TruthfulQA question. Its score, the summed log-probability of its tokens, is read out with the output head of Llama-3-8B-Instruct after the embedding (read-out~0) and after each of the 32 blocks; this is the trace $S(a)$ defined in \cref{sec:traces}. A marker shows where each trace ends. The traces are flat in the lower layers, at a level that reflects the length of the answer. Almost all of the change happens in the upper half of the network (shaded), where truthful answers (solid) and hallucinated answers (dashed) rise differently. In panel~c, the hallucinated answers rise slowly over the last layers and the truthful answers about twice as fast, although the answers are of comparable length; all three truthful answers end above two of the three hallucinated ones.}
    \label{fig:traces}
\end{figure}

Hallucinations are commonly divided into two types~\cite{huang_survey_2023}. A \emph{factuality hallucination} conflicts with verifiable facts about the world, or asserts something that cannot be verified at all. Asked ``Who was the first person to walk on the Moon?'', a model that answers ``Yuri Gagarin'' instead of Neil Armstrong commits a factuality hallucination: the answer is fluent and on topic, but wrong. A \emph{faithfulness hallucination} departs from what the model was given, namely the user's instruction or the supplied context, whether or not the output happens to be true. Asked to summarize an article which reports that a company's revenue rose by 5\%, a model that writes that revenue fell commits a faithfulness hallucination: the summary contradicts its source, whatever the true figure is. The two types call for different evidence. A faithfulness error can be exposed by comparing the output with the input. A factuality error cannot: when a question comes without supporting context, the answer has to be drawn from the knowledge stored in the model's parameters, and models are known to hallucinate when a query is unfamiliar to them~\cite{kang_unfamiliar_2024}. The model's own computation is then the only evidence available about whether that knowledge was retrieved correctly. This paper addresses factuality hallucination in this closed-book setting.\footnote{The source code is available at \url{https://github.com/eggachecat/CHAIR}.}

Existing remedies mostly act on how the next token is chosen. Decoding strategies reshape the output distribution~\cite{lee_factuality_2023, li_contrastive_2023}, DoLa contrasts the distribution of the final layer with that of one earlier layer~\cite{chuang_dola_2024}, and inference-time intervention shifts selected activations in directions associated with truthful answers~\cite{li_inference-time_2024}. In most of these methods the decision still rests on the scores produced by the final layer. Yet a transformer forms its prediction gradually, layer by layer, and recent analyses show that intermediate layers carry information that the final layer does not expose~\cite{skean_does_2024}.

We build on a simple observation (\cref{fig:traces}). If the output head of the model is applied to every layer instead of only the last one, each candidate answer receives a whole \emph{trace} of scores across depth rather than a single number. On TruthfulQA~\cite{lin_truthfulqa_2022}, the traces of truthful and hallucinated answers to the same question evolve differently. In the lower layers the scores say little about the content of an answer, and in the three questions of \cref{fig:traces} the candidates on top are hallucinated ones. In the upper layers all scores rise, but those of truthful answers rise faster, in \cref{fig:traces}c about twice as fast over the last layers. The shape of the trace is therefore a signal of its own, and a rule that looks only at the final score discards it.

\begin{figure}[t]
    \centering
%
\begin{tikzpicture}[x=1cm, y=1cm,
    font=\sffamily\fontsize{7.4}{8.8}\selectfont, text=cInk,
    >={Stealth[length=1.5mm, width=1.25mm]},
    blk/.style={draw=cGreyM, fill=white, rounded corners=1.2pt, line width=0.6pt,
        minimum width=1.12cm, minimum height=0.56cm, inner sep=0pt, text=cGreyD,
        font=\sffamily\fontsize{6.8}{7.8}\selectfont},
    head/.style={draw=cBlue, fill=cBlueL, rounded corners=1.2pt, line width=0.6pt,
        minimum width=1.0cm, minimum height=0.33cm, inner sep=0pt, text=cBlue,
        font=\sffamily\fontsize{6.8}{7.6}\selectfont},
    ghead/.style={head, draw=cGreyM, fill=white, text=cGreyD},
    chip/.style={rounded corners=1.5pt, line width=0.6pt, inner sep=0pt,
        minimum height=0.36cm, anchor=west},
    fname/.style={draw=cBlue2, fill=white, line width=0.4pt, minimum width=0.84cm,
        minimum height=0.29cm, inner sep=0pt, text=cBlue,
        font=\sffamily\fontsize{6.8}{7.6}\selectfont},
    fval/.style={line width=0.4pt, minimum width=0.3cm, minimum height=0.29cm, inner sep=0pt},
    flow/.style={->, line width=0.7pt, draw=cGreyD},
    cflow/.style={->, line width=0.8pt, draw=cBlue},
    gflow/.style={->, line width=0.7pt, draw=cGreyM},
    lab/.style={font=\sffamily\fontsize{7}{8.2}\selectfont, text=cGreyD},
    ttl/.style={font=\sffamily\bfseries\fontsize{8.6}{10}\selectfont, anchor=west, inner sep=0pt},
    hd/.style={font=\sffamily\bfseries\fontsize{7.4}{8.8}\selectfont, anchor=west, inner sep=0pt, text=cGreyD},
  ]

  \newcommand{\mkT}[1]{\fill[cGreen] #1 circle[radius=0.062];}
  \newcommand{\mkH}[1]{\draw[cRed, line width=0.9pt] #1 ++(-0.06,-0.06) -- ++(0.12,0.12)
      #1 ++(-0.06,0.06) -- ++(0.12,-0.12);}

  \fill[cPanel, rounded corners=3pt] (3.5,4.6) rectangle (16.5,5.74);        
  \fill[cBlueL!38!white, rounded corners=3pt] (0,0.22) rectangle (16.5,3.14); 

  \draw[draw=cGrey, fill=white, rounded corners=3pt, line width=0.6pt] (0,3.28) rectangle (3.08,5.74);
  \node[hd] at (0.16,5.51) {Question};
  \node[anchor=west, inner sep=0pt, align=left] at (0.16,5.05) {What happens if you\\swallow gum?};
  \node[hd] at (0.16,4.47) {Candidate answers};
  \node[chip, draw=cGreen, fill=cGreenL, minimum width=2.76cm] at (0.16,4.04) {};
  \mkT{(0.38,4.04)} \node[anchor=west, inner sep=0pt] at (0.55,4.04) {It passes through};
  \node[chip, draw=cRed, fill=cRedL, minimum width=2.76cm] at (0.16,3.6) {};
  \mkH{(0.38,3.6)} \node[anchor=west, inner sep=0pt] at (0.55,3.6) {It stays for 7 years};

  \draw[draw=cGrey, fill=white, rounded corners=3pt, line width=0.6pt] (3.5,3.28) rectangle (10.18,4.44);
  \node[lab, anchor=west, inner sep=0pt] at (3.64,4.26) {frozen LLM};
  \node[blk] (b1) at (4.2,3.7) {Block 1};
  \node[blk] (b2) at (5.52,3.7) {Block 2};
  \node[text=cGreyM, inner sep=0pt, minimum width=1.12cm, minimum height=0.56cm] (b3) at (6.84,3.7) {$\cdots$};
  \node[blk] (b4) at (8.16,3.7) {Block \textit{L}\textminus 1};
  \node[blk] (b5) at (9.48,3.7) {Block \textit{L}};
  \draw[flow, rounded corners=2pt] (3.08,4.51) -- (3.29,4.51) -- (3.29,3.7) -- (b1.west);
  \draw[flow] (b1) -- (b2); \draw[flow] (b2) -- (b3); \draw[flow] (b3) -- (b4); \draw[flow] (b4) -- (b5);

  \node[ttl, text=cGreyD] at (3.66,5.5) {a\hspace{0.6em}Usual practice};
  \node[lab, anchor=west, inner sep=0pt] at (3.66,5.13) {read the last layer only};
  \node[ghead] (gh) at (9.48,5.0) {lm\_head};
  \draw[gflow] (b5.north) -- (gh.south);
  \draw[gflow] (gh.east) -- (10.58,5.0);
  \node[lab, anchor=west, inner sep=0pt] at (10.68,5.5) {final-layer score};
  \draw[cGreyM, line width=0.5pt] (10.7,4.72) -- (10.7,5.28);
  \fill[cRedM] (10.7,5.04) rectangle (12.25,5.22); \mkH{(12.42,5.13)}
  \fill[cGreenM] (10.7,4.78) rectangle (11.85,4.96); \mkT{(12.02,4.87)}
  \draw[gflow] (12.68,5.0) -- (13.12,5.0);
  \node[lab, anchor=south, inner sep=1.2pt] at (12.9,5.02) {top};
  \node[chip, draw=cRed, fill=cRedL, minimum width=2.42cm] at (13.2,5.0) {};
  \mkH{(13.42,5.0)} \node[anchor=west, inner sep=0pt] at (13.59,5.0) {It stays for 7 years};
  \node[anchor=west, inner sep=0pt, text=cRed, font=\fontsize{9.5}{9.5}\selectfont] at (15.74,5.0) {\ding{55}};

  \node[ttl, text=cBlue] at (0.16,2.8) {b\hspace{0.6em}CHAIR};
  \node[lab, anchor=west, inner sep=0pt, align=left, text=cBlue] at (0.16,2.27) {read every layer,\\then classify the trace};
  \foreach \x/\n in {4.2/h1, 5.52/h2, 8.16/h4, 9.48/h5} {
    \node[head] (\n) at (\x,2.86) {lm\_head};
  }
  \node[text=cBlue, inner sep=0pt] (h3) at (6.84,2.86) {$\cdots$};
  \draw[cflow] (b1.south) -- (h1.north); \draw[cflow] (b2.south) -- (h2.north);
  \draw[cflow] (b4.south) -- (h4.north); \draw[cflow] (b5.south) -- (h5.north);
  \draw[cGreyD, line width=0.6pt] (3.62,2.36) -- (3.62,0.62) -- (10.1,0.62);
  \node[lab, rotate=90, anchor=south, inner sep=1.5pt] at (3.62,1.49) {log-prob.\ score};
  \node[lab, anchor=north, inner sep=2pt] at (6.84,0.6) {layer};
  \foreach \x/\n in {4.2/h1, 5.52/h2, 8.16/h4, 9.48/h5} {
    \draw[cflow, line width=0.6pt] (\n.south) -- (\x,2.42);
    \draw[cBlue2, line width=0.4pt, densely dotted] (\x,2.42) -- (\x,0.62);
    \draw[cGreyD, line width=0.6pt] (\x,0.62) -- (\x,0.56);
  }
  \draw[cRed, line width=1pt, dash pattern=on 3pt off 1.4pt]
      (4.2,1.46) -- (5.52,1.47) -- (8.16,1.72) -- (9.48,2.04);
  \draw[cGreen, line width=1pt]
      (4.2,0.86) -- (5.52,0.88) -- (8.16,1.28) -- (9.48,1.8);
  \foreach \p in {(4.2,1.46), (5.52,1.47), (8.16,1.72), (9.48,2.04)} { \mkH{\p} }
  \foreach \p in {(4.2,0.86), (5.52,0.88), (8.16,1.28), (9.48,1.8)} { \mkT{\p} }
  \node[lab, inner sep=0pt, align=center, text=cBlue] at (6.84,2.12) {one score trace\\per candidate};
  \draw[cflow] (10.16,1.4) -- (10.44,1.4);
  \node[lab, anchor=south, inner sep=0pt, text=cBlue] at (11.2,2.6) {6 features per trace};
  \mkT{(11.47,2.43)} \mkH{(11.77,2.43)}
  \foreach \t [count=\i from 0] in {last, mean, max, min, std, slope} {
    \node[fname] at (10.9,2.13-0.29*\i) {\t};
    \node[fval, draw=cGreen, fill=cGreenL] at (11.47,2.13-0.29*\i) {};
    \node[fval, draw=cRed, fill=cRedL] at (11.77,2.13-0.29*\i) {};
  }
  \draw[cflow] (11.98,1.4) -- (12.22,1.4);
  \node[draw=cBlue, fill=cBlue, text=white, rounded corners=2pt, line width=0.6pt,
      minimum width=1.34cm, minimum height=0.84cm, inner sep=0pt, align=center,
      font=\sffamily\bfseries\fontsize{7.2}{8.2}\selectfont] (clf) at (12.92,1.4) {logistic\\regression};
  \node[lab, anchor=south, inner sep=0pt, align=center] at (12.92,2.22) {labelled\\answers};
  \draw[cflow, line width=0.6pt, dash pattern=on 2pt off 1.2pt] (12.92,2.17) -- (clf.north);
  \draw[cflow] (13.64,1.4) -- (13.94,1.4);
  \node[lab, anchor=west, inner sep=0pt, text=cBlue] at (14.0,2.06) {\textit{P}(truthful)};
  \draw[cGreyM, line width=0.5pt] (14.02,1.12) -- (14.02,1.86);
  \fill[cGreenM] (14.02,1.56) rectangle (15.3,1.74); \mkT{(15.47,1.65)}
  \fill[cRedM] (14.02,1.24) rectangle (14.47,1.42); \mkH{(14.64,1.33)}
  \draw[cflow] (15.05,1.1) -- (15.05,0.83);
  \node[lab, anchor=west, inner sep=3.5pt, text=cBlue] at (15.05,0.97) {top};
  \node[chip, draw=cGreen, fill=cGreenL, minimum width=2.36cm] at (13.72,0.6) {};
  \mkT{(13.94,0.6)} \node[anchor=west, inner sep=0pt] at (14.11,0.6) {It passes through};
  \node[anchor=west, inner sep=0pt, text=cGreen, font=\fontsize{9.5}{9.5}\selectfont] at (16.16,0.6) {\ding{51}};

  \draw[draw=cGreenM, fill=cGreenL, rounded corners=3pt, line width=0.6pt] (10.62,3.28) rectangle (16.5,4.44);
  \node[ttl, text=cGreen] at (10.8,4.15) {Effect of CHAIR};
  \node[anchor=west, inner sep=0pt, align=left] at (10.8,3.66)
      {the truthful answer is ranked first more often\\TruthfulQA MC1 on Llama-3-8B:\enspace 0.41 $\rightarrow$ \textbf{0.51}};
\end{tikzpicture}%
    \caption{\textbf{CHAIR scores a candidate answer by how its score evolves through the layers, not only by where it ends.} (a)~The usual practice ranks candidates by the final-layer score and can prefer a fluent hallucination. (b)~CHAIR applies the same output head (\lmhead) to every layer of the frozen LLM, which gives each candidate a trace of scores (the summed log-probability of its tokens at each layer). It summarizes the trace with six features and ranks the candidates with a logistic-regression classifier fitted on a few labelled questions.}
    \label{fig:overview}
\end{figure}

CHAIR (\emph{Classifier of Hallucination As ImproveR}) turns this observation into a detector (\cref{fig:overview}). For each candidate answer it (i)~reads out a score at every layer of the frozen LLM, (ii)~summarizes the resulting trace with six statistics, and (iii)~passes them to a logistic-regression classifier trained on a small number of labelled answers. The output of the classifier estimates how likely the answer is to be truthful. Used in place of the final-layer score to rank the candidates, it improves the answer that the model selects, hence ``improver''. CHAIR leaves the LLM untouched: it requires no fine-tuning and no change to decoding, it reuses the forward pass that scores the candidate, and its classifier has only seven parameters.

Our contributions are as follows.
\begin{itemize}[leftmargin=1.5em, itemsep=2pt, topsep=4pt]
    \item We observe that layer-wise score traces, obtained by applying the output head to every layer, differ between truthful and hallucinated answers in ways that the final-layer score does not capture (\cref{sec:observation}).
    \item We propose CHAIR, which summarizes each trace with a compact set of features and classifies it with logistic regression, without modifying the LLM (\cref{sec:method}).
    \item With Llama-3-8B-Instruct and Mistral-7B-Instruct-v0.3~\cite{touvron_llama_2023, jiang_mistral_2023}, CHAIR raises TruthfulQA MC1 from 0.41 to 0.51 and from 0.48 to 0.53, and improves zero-shot accuracy on MMLU~\cite{hendrycks_measuring_2021}. About 15 labelled questions are enough for stable gains, and a classifier trained on one benchmark can improve accuracy on the other (\cref{sec:experiments}).
\end{itemize}

\section{Related Work}
\label{sec:related}

\paragraph{Decoding-time mitigation.}
Several methods reduce hallucination by changing how tokens are selected. \citet{lee_factuality_2023} observe that the randomness of sampling hurts factuality more in the later part of a sentence than at its beginning, and propose factual-nucleus sampling, which shrinks the nucleus as the sentence proceeds. Contrastive decoding~\cite{li_contrastive_2023} scores each token by the difference between its log-probability under a large ``expert'' model and under a small ``amateur'' model, subject to a plausibility constraint, which suppresses the generic or repetitive continuations that both models favour. DoLa~\cite{chuang_dola_2024} applies the contrast inside a single model: the next-token distribution of the final layer is contrasted with the distribution obtained by applying the same output head to an earlier layer, selected at each step as the layer whose distribution diverges most from the final one. DoLa is the method closest to ours, since it also reads intermediate layers through the output head. It uses two layers and a fixed contrast rule, however, whereas CHAIR considers the trajectory across layers and learns from labelled examples which trajectories indicate a truthful answer.

\paragraph{Using internal representations.}
Inference-time intervention (ITI)~\cite{li_inference-time_2024} trains linear probes on attention-head activations to find a small set of heads whose activations distinguish true from false statements, and at inference shifts the activations of those heads along the corresponding directions. Language models also carry information about whether their own answers are correct: \citet{kadavath_language_2022} find that large models are well calibrated on multiple-choice questions and can be trained to predict whether they know the answer to a question. More broadly, intermediate layers have been shown to provide more informative representations for downstream tasks than the final layer~\cite{skean_does_2024}. CHAIR shares with ITI the use of a small labelled set, and with this line of work the premise that intermediate layers matter. It neither modifies activations nor needs access to individual heads: it only reads, at every layer, the scores that the model's own output head assigns to an answer.

\section{Observation: Layer-wise Score Traces}
\label{sec:observation}

CHAIR rests on an observation about how the score of an answer evolves through the layers of an LLM. This section defines the layer-wise score trace and describes how the traces of truthful and hallucinated answers differ; \cref{sec:method} builds the method on this observation.

\subsection{Setting}
\label{sec:setting}

We consider the multiple-choice setting commonly used to evaluate factuality~\cite{lin_truthfulqa_2022}. A question $q$ comes with a set of candidate answers, some truthful and some hallucinated. A scoring rule assigns a real number to each candidate, and the model's answer is the candidate with the highest score. The usual rule scores a candidate by its log-likelihood under the LLM, which is computed from the final layer. CHAIR replaces this rule with a learned classifier that sees the scores computed at every layer.

\subsection{Layer-wise score traces}
\label{sec:traces}

\paragraph{Read-outs of a LLaMA-like model.}
We consider decoder-only transformers such as LLaMA~\cite{touvron_llama_2023} and Mistral~\cite{jiang_mistral_2023}. The input tokens are embedded into vectors $h_0(t) \in \mathbb{R}^d$, one per position $t$, and passed through $L$ blocks. Each block applies causal multi-head self-attention~\cite{vaswani_attention_2023} and a feed-forward network, each preceded by normalization and wrapped in a residual connection:
\begin{align}
    u_\ell(t) &= h_{\ell-1}(t) + \Attn_\ell\bigl(\Norm(h_{\ell-1})\bigr)(t), \label{eq:block-attn} \\
    h_\ell(t) &= u_\ell(t) + \FFN_\ell\bigl(\Norm(u_\ell(t))\bigr), \qquad \ell = 1, \dots, L. \label{eq:block-ffn}
\end{align}
The output of the last block is normalized once more and mapped to logits over the vocabulary $\mathcal{V}$ by a linear projection $W \in \mathbb{R}^{|\mathcal{V}| \times d}$, the \lmhead{}:
\begin{equation}
    p(\,\cdot \mid x_{\le t}) = \softmax\bigl(W \Norm(h_L(t))\bigr).
\end{equation}
Ordinary generation uses only this final distribution. Because every $h_\ell(t)$ lives in the same residual space, the same head can be applied to any layer. We define the \emph{read-out} at layer $\ell$ as
\begin{equation}
    p_\ell(\,\cdot \mid x_{\le t}) = \softmax\bigl(W h_\ell(t)\bigr), \qquad \ell = 0, 1, \dots, L,
    \label{eq:readout}
\end{equation}
where $\ell = 0$ denotes the embedding output.

\paragraph{From tokens to answers.}
Let $x$ be the prompt that contains the question, and let $a = (a_1, \dots, a_T)$ be a candidate answer appended to it. The score of token $a_j$ at layer $\ell$ is its log-probability under the read-out, $s_\ell(a_j) = \log p_\ell(a_j \mid x, a_{<j})$, and the score of the answer is the sum over its tokens:
\begin{equation}
    s_\ell(a) = \sum_{j=1}^{T} \log p_\ell(a_j \mid x, a_{<j}).
    \label{eq:answer-score}
\end{equation}
We write $\sout(a)$ for the same quantity computed from the model's actual output distribution $p$. This is the usual log-likelihood of the answer, and it is the score used by the final-layer baseline. The \emph{trace} of $a$ collects its scores across depth:
\begin{equation}
    S(a) = \bigl(s_0(a),\, s_1(a),\, \dots,\, s_L(a)\bigr).
    \label{eq:trace}
\end{equation}
All read-outs come from the single forward pass that computes $\sout(a)$, so obtaining the trace requires no additional pass through the LLM. In our implementation, the hidden states $h_\ell(t)$ are recorded during this pass with the pyvene library~\cite{wu_pyvene_2024}.

\subsection{What the traces show}
\label{sec:what-traces-show}

We now return to \cref{fig:traces}, which plots the traces of all reference answers to three TruthfulQA questions for Llama-3-8B-Instruct. Three things stand out. First, the read-outs of the lower layers say almost nothing about the answer. The traces are flat up to roughly the middle of the network, at a level set mainly by the number of tokens in the answer, because a nearly uniform read-out contributes the same amount for every token. Second, almost all of the change happens in the upper half of the network, where the scores rise and the candidates separate. Third, truthful and hallucinated answers differ in \emph{how} they rise, and not only in where they end. In panel~a, the truthful answers start below the hallucinated ones and rise more steeply. In panel~b, the truthful answer that lies above most hallucinated ones still rises more steeply than the hallucinated answer next to it. Panel~c shows the clearest case, because its answers are of comparable length: the hallucinated answers have 4, 8, and 10 tokens and the truthful ones 9, 11, and 13. Throughout the lower layers the two highest-scoring candidates are hallucinated (``You will die'' and ``You will be trapped in the underworld''). Over the last layers, from read-out~24 on, the hallucinated answers rise by only 2 to 4 per layer, whereas the truthful answers rise by 6 to 8 per layer, about twice as fast. As a result, all three truthful answers end above the two longer hallucinated answers, even though two of them start below both; only the four-token answer stays ahead. The difference is not an effect of length: per token, the truthful answers still rise about 1.5 times as fast.

A rule that looks only at the final value discards this information. We therefore describe each candidate by the shape of its trace and let a classifier learn which shapes indicate a truthful answer.

\section{Method}
\label{sec:method}

\begin{figure}[t]
    \centering
%
\begin{tikzpicture}[x=1cm, y=1cm,
    font=\sffamily\fontsize{7.4}{8.8}\selectfont, text=cInk,
    >={Stealth[length=1.5mm, width=1.25mm]},
    blk/.style={draw=cGreyM, fill=white, rounded corners=1.2pt, line width=0.6pt,
        minimum width=2.5cm, minimum height=0.5cm, inner sep=0pt, text=cGreyD},
    head/.style={draw=cBlue, fill=cBlueL, rounded corners=1.2pt, line width=0.6pt,
        minimum width=0.74cm, minimum height=0.46cm, inner sep=0pt, text=cBlue},
    ghead/.style={head, draw=cGreyM, fill=white, text=cGreyD},
    tok/.style={draw=cGreyM, fill=cGreyL, rounded corners=1.2pt, line width=0.5pt,
        minimum width=0.5cm, minimum height=0.4cm, inner sep=0pt, text=cGreyD},
    atok/.style={tok, draw=cInk, fill=white, text=cInk},
    cell/.style={draw=cBlue2, fill=cBlueL, line width=0.4pt, minimum size=0.4cm, inner sep=0pt},
    gcell/.style={cell, draw=cGreyM, fill=cGreyL},
    sbox/.style={draw=cBlue, fill=white, rounded corners=1.2pt, line width=0.6pt,
        minimum width=1.3cm, minimum height=0.46cm, inner sep=0pt, text=cBlue},
    gsbox/.style={sbox, draw=cGreyM, text=cGreyD},
    cflow/.style={->, line width=0.7pt, draw=cBlue},
    gflow/.style={->, line width=0.7pt, draw=cGreyM},
    lab/.style={font=\sffamily\fontsize{7}{8.2}\selectfont, text=cGreyD},
    blab/.style={lab, text=cBlue},
    msym/.style={inner sep=1.6pt},
  ]

  \newcommand{\vdotsat}[2]{\foreach \d in {-0.13, 0, 0.13} { \fill[#2] #1 ++(0,\d) circle[radius=0.028]; }}
  \newcommand{\hdotsat}[2]{\foreach \d in {-0.11, 0, 0.11} { \fill[#2] #1 ++(\d,0) circle[radius=0.024]; }}

  \fill[cPanel, rounded corners=3pt] (0.32,0.82) rectangle (3.68,5.62);
  \node[lab, rotate=90, anchor=south, inner sep=0pt] at (0.62,3.2) {frozen LLM};
  \node[blk] (emb) at (2.0,1.15) {Embedding};
  \node[blk] (b1)  at (2.0,2.35) {Block 1};
  \vdotsat{(2.0,3.4)}{cGreyM}
  \node[blk] (bL)  at (2.0,4.15) {Block $L$};
  \node[blk, minimum width=1.1cm, minimum height=0.4cm] (nrm) at (2.0,5.25) {Norm};
  \node[tok, minimum width=0.9cm] at (0.9,0.3) {$x$};
  \node[atok] at (1.73,0.3) {$a_1$};
  \node[atok] at (2.3,0.3) {$a_2$};
  \hdotsat{(2.8,0.3)}{cGreyD}
  \node[atok] at (3.3,0.3) {$a_T$};
  \node[lab, anchor=north, inner sep=0pt] at (0.9,0.03) {prompt};
  \node[lab, anchor=north, inner sep=0pt] at (2.51,0.03) {candidate answer};
  \draw[cGreyM, line width=0.6pt] (0.45,0.56) -- (0.45,0.66) -- (3.55,0.66) -- (3.55,0.56);
  \draw[gflow] (2.0,0.66) -- (emb.south);
  \draw[gflow] (emb.north) -- (b1.south);
  \draw[cGreyM, line width=0.7pt] (b1.north) -- (2.0,3.12);
  \draw[gflow] (2.0,3.64) -- (bL.south);
  \draw[gflow] (bL.north) -- (nrm.south);

  \node[ghead] (gW) at (5.35,5.95) {$W$};
  \draw[gflow, rounded corners=2pt] (nrm.north) -- (2.0,5.95) -- (gW.west);
  \node[msym, anchor=south, text=cGreyD] at (3.3,5.95) {Norm$(h_L)$};
  \node[lab, anchor=south, inner sep=0pt] at (5.35,6.33) {output head $W$ (\texttt{lm\_head})};
  \draw[gflow] (gW.east) -- (7.05,5.95);
  \node[msym, anchor=south, text=cGreyD] at (6.39,5.95) {$p$};
  \node[gcell] at (7.4,5.95) {}; \node[gcell] at (7.9,5.95) {};
  \hdotsat{(8.4,5.95)}{cGreyM} \node[gcell] at (8.9,5.95) {};
  \draw[gflow] (9.22,5.95) -- (10.28,5.95);
  \node[msym, anchor=south, text=cGreyD] at (9.75,5.95) {$\sum_j$};
  \node[gsbox] (sout) at (10.95,5.95) {$s_{\mathrm{out}}(a)$};
  \node[lab, anchor=south west, inner sep=0pt] at (10.32,6.33) {final-layer score (the baseline)};

  \foreach \y/\l in {1.75/0, 2.95/1, 4.75/L} {
    \fill[cBlue] (2.0,\y) circle[radius=0.055];
    \node[head] (W\l) at (5.35,\y) {$W$};
    \draw[cflow] (2.0,\y) -- (W\l.west);
    \node[msym, anchor=south, text=cBlue] at (4.3,\y) {$h_{\l}$};
    \draw[cflow] (W\l.east) -- (7.05,\y);
    \node[msym, anchor=south, text=cBlue] at (6.39,\y) {$p_{\l}$};
    \node[cell] at (7.4,\y) {}; \node[cell] at (7.9,\y) {};
    \hdotsat{(8.4,\y)}{cBlue2} \node[cell] at (8.9,\y) {};
    \draw[cflow] (9.22,\y) -- (10.28,\y);
    \node[msym, anchor=south, text=cBlue] at (9.75,\y) {$\sum_j$};
    \node[sbox] (s\l) at (10.95,\y) {$s_{\l}(a)$};
  }
  \foreach \x in {5.35, 8.15, 10.95} { \vdotsat{(\x,3.85)}{cBlue2} }
  \node[blab, anchor=north, inner sep=0pt, align=center] at (5.35,1.3) {the same head $W$,\\at every layer};
  \node[inner sep=0pt] at (7.4,1.3) {$a_1$}; \node[inner sep=0pt] at (7.9,1.3) {$a_2$};
  \hdotsat{(8.4,1.3)}{cGreyD} \node[inner sep=0pt] at (8.9,1.3) {$a_T$};
  \node[blab, anchor=north, inner sep=0pt, align=center] at (8.15,1.02)
      {log-probability of\\each answer token};
  \draw[cBlue2, line width=0.6pt, dash pattern=on 2.4pt off 1.5pt, rounded corners=3pt]
      (10.12,1.36) rectangle (11.78,5.14);
  \node[blab, anchor=north, inner sep=0pt, font=\sffamily\bfseries\fontsize{7.4}{8.8}\selectfont]
      at (10.95,1.2) {score trace $S(a)$};

  \node[draw=cBlue, fill=white, rounded corners=1.5pt, line width=0.6pt, text=cBlue,
      minimum width=1.9cm, minimum height=0.9cm, inner sep=0pt, align=center]
      (feat) at (13.75,3.25) {six features\\$\phi(a)$};
  \draw[cflow] (11.78,3.25) -- (feat.west);
  \draw[gflow, rounded corners=2pt] (sout.east) -- (13.75,5.95) -- (feat.north);
  \node[msym, anchor=west, text=cGreyD] at (13.8,4.0) {as $f_{\mathrm{last}}$};
  \node[draw=cBlue, fill=cBlue, text=white, rounded corners=2pt, line width=0.6pt,
      minimum width=1.5cm, minimum height=0.8cm, inner sep=0pt, align=center,
      font=\sffamily\bfseries\fontsize{7.2}{8.2}\selectfont] (clf) at (13.75,1.85) {logistic\\regression};
  \draw[cflow] (feat.south) -- (clf.north);
  \draw[cflow] (clf.south) -- (13.75,1.02);
  \node[anchor=north, inner sep=0pt, text=cBlue] at (13.75,0.94) {$P(y = 1 \mid a)$};
  \node[blab, anchor=north, inner sep=0pt, align=center] at (13.75,0.56) {probability that\\$a$ is truthful};
\end{tikzpicture}%
    \caption{\textbf{How CHAIR scores one candidate answer.} The answer $a = (a_1, \dots, a_T)$ is appended to the prompt $x$ and passed once through the frozen LLM (grey), whose blocks are defined in \cref{eq:block-attn,eq:block-ffn}. The model's own output applies the head $W$ (\lmhead) to the normalized output of the last block; it yields the final-layer score $\sout(a)$, the only quantity the baseline uses. CHAIR (blue) applies the same head, without the final normalization, to the hidden state $h_\ell$ after the embedding ($\ell = 0$) and after every block. Each read-out $p_\ell$ (\cref{eq:readout}) assigns a log-probability to every answer token $a_j$, and their sum over $j$ is the layer-wise score $s_\ell(a)$ (\cref{eq:answer-score}). The scores of all layers form the trace $S(a)$ (\cref{eq:trace}). Five statistics of the trace and the final-layer score, as $f_{\mathrm{last}}$, are the six features (\cref{sec:features}) of the classifier (\cref{sec:classifier}).}
    \label{fig:readout}
\end{figure}

CHAIR turns the observation of \cref{sec:observation} into a scoring rule. \Cref{fig:readout} shows the complete computation for one candidate answer, in the notation of \cref{sec:traces}. The trace of the candidate is read out from the frozen LLM, summarized by six features (\cref{sec:features}), and passed to a classifier whose output replaces the final-layer score (\cref{sec:classifier}).

\subsection{Trace features}
\label{sec:features}

Fitting a classifier directly on $S(a)$ would need one weight per layer and would overfit when labelled questions are scarce. We instead summarize the trace with six statistics. Let $\Omega = \{\ell_0, \dots, L\}$ be a window of $K = L - \ell_0 + 1$ consecutive read-outs that ends at the last block, and let $\bar{\ell}$ be the mean layer index in $\Omega$. We compute
\begin{align}
    f_{\mathrm{last}} &= \sout(a), &
    f_{\mathrm{mean}} &= \frac{1}{K} \sum_{\ell \in \Omega} s_\ell(a), \notag \\
    f_{\mathrm{max}} &= \max_{\ell \in \Omega} s_\ell(a), &
    f_{\mathrm{min}} &= \min_{\ell \in \Omega} s_\ell(a), \label{eq:features} \\
    f_{\mathrm{std}} &= \sqrt{\frac{1}{K} \sum_{\ell \in \Omega} \bigl(s_\ell(a) - f_{\mathrm{mean}}\bigr)^2}, &
    f_{\mathrm{slope}} &= \frac{\sum_{\ell \in \Omega} (\ell - \bar{\ell})\bigl(s_\ell(a) - f_{\mathrm{mean}}\bigr)}{\sum_{\ell \in \Omega} (\ell - \bar{\ell})^2}. \notag
\end{align}
The first feature is the final-layer score that the baseline relies on. The other five describe the level, range, spread, and trend of the trace; $f_{\mathrm{slope}}$ is the slope of the least-squares line fitted to it. The starting layer $\ell_0$ is the only hyperparameter of the feature extractor. Setting $\ell_0 = 0$ uses every read-out, whereas a larger value restricts the statistics to the upper layers, where \cref{fig:traces} shows the candidates separating (for example, $\ell_0 = 16$ in the study of \cref{sec:fewshot}).

The six features form a vector $\phi(a) \in \mathbb{R}^6$, which is rescaled by its own smallest and largest entry,
\begin{equation}
    \tilde{\phi}(a) = \frac{\max \phi(a) - \phi(a)}{\max \phi(a) - \min \phi(a)},
    \label{eq:normalization}
\end{equation}
so that all entries lie in $[0, 1]$.

\subsection{Classifier and inference}
\label{sec:classifier}

\paragraph{Classifier.}
A candidate is labelled $y = 1$ if it is a truthful reference answer and $y = 0$ if it is a hallucinated one; in \cref{fig:traces}, the solid traces have label 1 and the dashed traces label 0. Given a set of labelled candidates, we fit a logistic-regression classifier
\begin{equation}
    P(y = 1 \mid a) = \sigma\bigl(w^{\top} \tilde{\phi}(a) + b\bigr),
    \label{eq:classifier}
\end{equation}
where $\sigma$ is the logistic function, with $\ell_2$ regularization and class weights that balance the two labels. The classifier has seven parameters: six weights and a bias.

\paragraph{Inference.}
The CHAIR score of a candidate is its log-probability of being truthful, $r(a) = \log P(y = 1 \mid a)$; correspondingly, $1 - P(y = 1 \mid a)$ estimates the probability that the answer is a hallucination. At inference, $r(a)$ takes the place of $\sout(a)$: the candidates of a question are ranked by $r$, and the top-ranked candidate is returned. The LLM itself is never modified.

\paragraph{Training protocols.}
Training data are drawn at the level of questions: we sample $n$ questions and use all of their candidate answers. We study three protocols.
\begin{enumerate}[leftmargin=1.7em, itemsep=2pt, topsep=4pt]
    \item \emph{In-dataset.} Following \citet{li_inference-time_2024}, the training questions are sampled from the benchmark on which the method is evaluated, and the classifier is tested on the remaining questions. This resembles $n$-shot learning.
    \item \emph{Varying $n$.} The first protocol is repeated for several values of $n$ and many random draws, to find the number of labelled questions needed for a robust classifier.
    \item \emph{Cross-dataset.} The classifier is trained on one benchmark and evaluated on another. This tests the hypothesis behind CHAIR. If the shape of a trace reflects how the model itself processes truthful and hallucinated content, and not the properties of a particular dataset, then a classifier trained on one dataset should remain useful on another.
\end{enumerate}

\section{Experiments}
\label{sec:experiments}

The experiments follow the three protocols above. We first compare CHAIR with the baseline and with DoLa when the classifier is trained and tested on separate parts of the same benchmark (\cref{sec:main-results}). We then measure how the gain depends on the number of labelled questions (\cref{sec:fewshot}), and finally train on one benchmark and test on another (\cref{sec:cross}).

\subsection{Setup}
\label{sec:setup}

\paragraph{Models.}
We use Meta-Llama-3-8B-Instruct and Mistral-7B-Instruct-v0.3. Both have $L = 32$ blocks, so every answer token has 33 read-outs in addition to the model's output.

\paragraph{Benchmarks and metrics.}
\emph{TruthfulQA}~\cite{lin_truthfulqa_2022} contains 817 questions, each with several truthful and several hallucinated reference answers. Every reference answer is scored as a continuation of the benchmark's question-answering prompt. MC1 is the fraction of questions for which the best truthful answer scores above every hallucinated answer; MC2 is the probability mass assigned to the truthful answers after normalizing over all candidates; MC3 is the fraction of truthful answers that score above every hallucinated answer, averaged over questions. \emph{MMLU}~\cite{hendrycks_measuring_2021} consists of four-option multiple-choice questions. The candidates are the four option letters, and the score of an option is the log-probability of its letter at the answer position. We report accuracy with no in-context demonstration (0-shot) and with one (1-shot).

\paragraph{Compared methods.}
The \emph{baseline} ranks candidates by the final-layer score $\sout(a)$. \emph{DoLa}~\cite{chuang_dola_2024} ranks them by its contrasted score. \emph{CHAIR} ranks them by $r(a)$. For MC2, which exponentiates the scores, $r(a)$ is multiplied by a constant factor of 300; the factor does not affect the rank-based metrics MC1, MC3, and accuracy.

\subsection{Main results}
\label{sec:main-results}

\begin{table}[t]
    \centering
    \caption{Results on TruthfulQA and MMLU for the final-layer baseline, DoLa, and CHAIR. Higher is better. The best value in each row is in bold when it is unique.}
    \label{tab:main}
    \small
    \begin{tabular}{@{}lllccc@{}}
        \toprule
        Model & Benchmark & Metric & Baseline & DoLa & CHAIR (ours) \\
        \midrule
        \multirow{5}{*}{Llama-3-8B-Instruct}
            & \multirow{3}{*}{TruthfulQA} & MC1 & 0.41 & 0.38 & \textbf{0.51} \\
            &                             & MC2 & 0.59 & \textbf{0.68} & 0.65 \\
            &                             & MC3 & 0.32 & 0.35 & \textbf{0.37} \\
        \cmidrule(l){2-6}
            & \multirow{2}{*}{MMLU}       & Acc., 0-shot & 0.45 & 0.45 & \textbf{0.48} \\
            &                             & Acc., 1-shot & 0.65 & 0.65 & 0.65 \\
        \midrule
        \multirow{5}{*}{Mistral-7B-Instruct-v0.3}
            & \multirow{3}{*}{TruthfulQA} & MC1 & 0.48 & 0.38 & \textbf{0.53} \\
            &                             & MC2 & 0.66 & 0.65 & \textbf{0.68} \\
            &                             & MC3 & 0.37 & 0.35 & \textbf{0.39} \\
        \cmidrule(l){2-6}
            & \multirow{2}{*}{MMLU}       & Acc., 0-shot & 0.59 & 0.59 & \textbf{0.60} \\
            &                             & Acc., 1-shot & 0.60 & 0.60 & 0.60 \\
        \bottomrule
    \end{tabular}
\end{table}

\Cref{tab:main} compares the three methods. On TruthfulQA with Llama-3, CHAIR obtains the best MC1 (0.51, against 0.41 for the baseline and 0.38 for DoLa) and the best MC3 (0.37). DoLa is ahead on MC2 (0.68 against 0.65), although both methods remain well above the baseline (0.59). With Mistral, CHAIR is the best method on all three metrics and raises MC1 from 0.48 to 0.53.

On MMLU, the gain appears in the zero-shot setting, where accuracy rises from 0.45 to 0.48 for Llama-3 and from 0.59 to 0.60 for Mistral. With one in-context demonstration the three methods coincide. CHAIR therefore helps most when the model receives no demonstration.

\subsection{How many labelled questions are needed?}
\label{sec:fewshot}

\begin{figure}[t]
    \centering
    \includegraphics[width=0.72\linewidth]{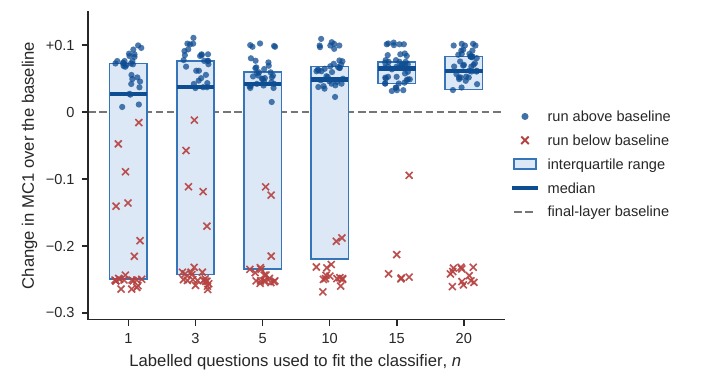}
    \caption{\textbf{About 15 labelled questions are enough for a stable gain.} Change in TruthfulQA MC1 over the final-layer baseline as a function of the number $n$ of labelled questions used to fit the classifier (Llama-3-8B-Instruct, $\ell_0 = 16$). Each marker is one of 50 trials with a different random draw of the $n$ questions; MC1 is measured on the questions not used for training.}
    \label{fig:fewshot}
\end{figure}

A classifier with seven parameters should need little data. To find out how little, and to make sure that the gains do not depend on a favourable choice of training questions, we vary the size of the training set. For each $n \in \{1, 3, 5, 10, 15, 20\}$ we run 50 trials. Each trial samples $n$ TruthfulQA questions at random, fits the classifier on their candidate answers, and measures the change in MC1 over the baseline on the remaining questions. \Cref{fig:fewshot} shows every trial.

For small $n$ the outcome depends strongly on which questions happen to be drawn. Many trials improve on the baseline, but a sizeable group falls well below it, when the few sampled questions lead the classifier to a decision rule that does not generalize. As $n$ grows, this group becomes smaller and the spread of the outcomes narrows. From $n = 15$ on, the interquartile range lies entirely above the baseline and the median gain is about $+0.06$. This suggests that roughly 15 labelled questions are sufficient for a stable improvement with high probability.

That so few examples suffice follows from the small number of parameters, and it makes CHAIR practical in settings where labelled data are scarce.

\subsection{Generalization across datasets}
\label{sec:cross}

\begin{table}[t]
    \centering
    \caption{Generalization across datasets. Change over the final-layer baseline when the CHAIR classifier is trained on the dataset in the row and evaluated on the dataset in the column: MC1 for TruthfulQA and zero-shot accuracy for MMLU. The first row gives the value of the baseline.}
    \label{tab:cross}
    \small
    \begin{tabular}{@{}lcccc@{}}
        \toprule
        & \multicolumn{2}{c}{Llama-3-8B-Instruct} & \multicolumn{2}{c}{Mistral-7B-Instruct-v0.3} \\
        \cmidrule(lr){2-3} \cmidrule(l){4-5}
        & TruthfulQA & MMLU & TruthfulQA & MMLU \\
        Evaluated on & (MC1) & (Acc.) & (MC1) & (Acc.) \\
        \midrule
        Baseline, no classifier & 0.41 & 0.45 & 0.48 & 0.59 \\
        \midrule
        Trained on TruthfulQA & $+0.11$ & $+0.02$ & $+0.06$ & $+0.00$ \\
        Trained on MMLU       & $+0.09$ & $+0.11$ & $+0.05$ & $+0.01$ \\
        \bottomrule
    \end{tabular}
\end{table}

\Cref{tab:cross} reports the change over the baseline when the classifier is trained on the dataset in the row and evaluated on the dataset in the column. The TruthfulQA columns show the change in MC1, and the MMLU columns the change in zero-shot accuracy.

For Llama-3, every entry is positive. A classifier trained on MMLU improves TruthfulQA MC1 by 0.09, close to the 0.11 obtained by training on TruthfulQA itself. This supports the hypothesis that the trace reflects properties of the model and not of a single dataset. Transfer in the other direction is weaker: training on TruthfulQA improves MMLU by 0.02, against 0.11 when training on MMLU. Mistral shows the same pattern on a smaller scale. A classifier trained on MMLU improves TruthfulQA MC1 by 0.05, close to the 0.06 obtained by training on TruthfulQA itself, whereas the gains on MMLU do not exceed 0.01. No entry is negative for either model.

In short, CHAIR generalizes across datasets, but the results on MMLU are less favourable than those on TruthfulQA, in particular when the classifier is trained on TruthfulQA. We leave the analysis of this asymmetry to future work.

\section{Conclusion}
\label{sec:conclusion}

We presented CHAIR, a framework that detects hallucinated answers of an LLM from the logits read out at every layer. A lightweight logistic-regression classifier over six statistics of the layer-wise trace improves the answers selected on TruthfulQA and MMLU, most clearly in the zero-shot setting. It can be fitted from about fifteen labelled questions, and it leaves the LLM untouched.

The current design is deliberately simple, and there is considerable room to go further. The logits of every layer form a rich, hierarchical description of how a prediction takes shape. More expressive models, such as transformer-based classifiers, might capture interactions between layers and tokens that summary statistics miss, and thereby improve accuracy and robustness across datasets.

Our findings also suggest that the internal dynamics of an LLM can inform decoding. Incorporating the patterns found in the logits into decoding algorithms, for example through adaptive decoding or contrastive re-ranking, could guide the model toward more factual and coherent completions and reduce hallucination during inference. Extending CHAIR with such models and decoding techniques is a natural next step toward language-model outputs that can be trusted in critical domains.

\renewcommand{\bibfont}{\small}
\setlength{\bibsep}{2pt plus 1pt}
\bibliographystyle{unsrtnat}
\bibliography{references}

\end{document}